%% file: main.tex
\documentclass[11pt,letterpaper]{article}

\newif\ifnips\nipsfalse
\ifnips
\usepackage{nips_2018}
\else
\usepackage[margin=1in]{geometry}
\fi

\input{Commands}

\input{libtheorems.tex}

\usepackage[hidelinks]{hyperref}
\usepackage{url}
\usepackage[numbers]{natbib}
\usepackage{graphicx,subfigure,amsmath,amssymb,amsfonts,bm,epsfig,epsf,url,dsfont,color}

\usepackage{amsthm}
\newtheorem{theorem}{Theorem}[section]

\let\P\undefined
\DeclareMathOperator*{\P}{\mathbb{P}}

\title{Adversarial Examples from Cryptographic\\ Pseudo-Random Generators}

\ifnips\else\def\And{\and}\fi
\author{S\'ebastien Bubeck\\
       Microsoft Research\\
			 \texttt{sebubeck@microsoft.com}\\
			 \And
Yin Tat Lee\\
       University of Washington \&\\
       Microsoft Research\\
			 \texttt{yintat@uw.edu}\\
			 \And
       Eric Price\\
       UT Austin\\
			 \texttt{ecprice@cs.utexas.edu}\\
			 \And
      Ilya Razenshteyn\\
       Microsoft Research\\
			 \texttt{ilyaraz@microsoft.com}\\
}

\begin{document}

\maketitle

\begin{abstract}
In our recent work (Bubeck, Price, Razenshteyn, arXiv:1805.10204) we argued that adversarial examples in machine learning might be due to an inherent computational hardness of the problem. More precisely, we constructed a binary classification task for which (i) a robust classifier exists; yet no non-trivial accuracy can be obtained with an efficient algorithm in (ii) the statistical query model. In the present paper we significantly strengthen both (i) and (ii): we now construct a task which admits (i') a {\em maximally} robust classifier (that is it can tolerate perturbations of size comparable to the size of the examples themselves); and moreover we prove computational hardness of learning this task under (ii') a standard cryptographic assumption.
\end{abstract}

\input{intro.tex}
\input{addendum.tex}
\input{proof.tex}
\bibliographystyle{plainnat}
\bibliography{bib}

\end{document}

%% file: Commands.tex
\renewcommand{\phi}{\varphi}

\newcommand{\N}{\mathbb{N}}
\newcommand{\R}{\mathbb{R}}

\newcommand{\cD}{\mathcal{D}}

\def\ds1{\mathds{1}}
\renewcommand{\epsilon}{\varepsilon}
\newcommand{\eps}{\epsilon}

\newlength{\minipagewidth}
\newcommand{\beq}{\begin{equation}}
\newcommand{\eeq}{\end{equation}}

\newcommand{\beqa}{\begin{eqnarray}}
\newcommand{\eeqa}{\end{eqnarray}}

\newcommand{\beqan}{\begin{eqnarray*}}
\newcommand{\eeqan}{\end{eqnarray*}}

\def\ba#1\ea{\begin{align*}#1\end{align*}} 
\def\banum#1\eanum{\begin{align}#1\end{align}} 



%% file: libtheorems.tex
\usepackage{etoolbox}

\makeatletter

\newcommand{\define}[4][ignore]{%
  \ifstrequal{#1}{ignore}{}{
  \@namedef{thmtitle@#2}{#1}}%
  \@namedef{thm@#2}{#4}%
  \@namedef{thmtypen@#2}{lemma}%
  \newtheorem{thmtype@#2}[theorem]{#3}%
  \newtheorem*{thmtypealt@#2}{#3~\ref{#2}}%
}

\newcommand{\state}[1]{%
  \@namedef{curthm}{#1}
  \@ifundefined{thmtitle@#1}{
  \begin{thmtype@#1}
    }{
  \begin{thmtype@#1}[\@nameuse{thmtitle@#1}]
  }
    \label{#1}
    \@nameuse{thm@#1}
  \end{thmtype@#1}
  \@ifundefined{thmdone@#1}{
  \@namedef{thmdone@#1}{stated}%
  }{}
}

\newcommand{\restate}[1]{%
  \@namedef{curthm}{#1}
  \@ifundefined{thmtitle@#1}{
    \begin{thmtypealt@#1}
    }{
  \begin{thmtypealt@#1}[\@nameuse{thmtitle@#1}]
  }
    \@nameuse{thm@#1}
  \end{thmtypealt@#1}
  \@ifundefined{thmdone@#1}{
  \@namedef{thmdone@#1}{stated}%
  }{}
}

\newcommand{\thmlabel}[1]{
  \@ifundefined{thmdone@\@nameuse{curthm}}{\label{#1}
    }{\tag*{\eqref{#1}}}
}
\makeatother

%% file: intro.tex
\section{Introduction}
We refer to our prior work \cite{BPR18} (henceforth referred to as BPR) for context and references about the emerging field of robust machine learning. Our primary goal in the present paper is to describe in the simplest possible way our new conceptual contribution, namely that even assuming the existence of a {\em maximally} robust classifier one cannot hope in general to attain \emph{efficiently} any kind of robustness whatsoever, and that this impossibility result can be obtained from a widely-accepted computational hardness assumption.

Let us start by briefly recalling the general setting and the BPR result. Basic binary classification can be phrased as follows: given $\mathrm{poly}(n)$ i.i.d.\ samples from two distributions $\cD_0$ and $\cD_1$ supported on $\R^n$, find (if possible in $\mathrm{poly}(n)$ time) a set $A \subset \R^n$ such that
\begin{equation} \label{eq:nonrobust}
\P_{X \sim \cD_0}(X \in A) \geq 0.99 \,, \text{ and } \P_{X \sim \cD_1}(X \not\in A) \geq 0.99 \,.
\end{equation}
The {\em $\epsilon$-robust} version asks for the following more stringent requirement:
\begin{equation} \label{eq:robust}
\P_{X \sim \cD_0}(B(X,\epsilon) \in A) \geq 0.99 \,, \text{ and } \P_{X \sim \cD_1}(B(X,\epsilon) \not\in A) \geq 0.99 \,,
\end{equation}
where $B(x,\epsilon) = \{ z \in \R^n : \|z - x\| \leq \epsilon\}$. In this paper we restrict our attention to the Euclidean norm case (however the same results hold say for $\ell_{\infty}$ norm). We refer to the set $A$ as a {\em classifier}, and we say that it is a {\em maximally robust} classifier if it satisfies \eqref{eq:robust} with $\epsilon = \Theta(\max_{i \in \{0,1\}} \mathrm{diam}(\mathrm{supp}(\cD_i))$.

Current state of the art machine learning essentially fails at the robust task \eqref{eq:robust} for large-scale problems. BPR identified four mutually exclusive possibilities to explain this current state of affair:
\begin{enumerate}
\item No robust classifier exists.
\item Identifying a robust classifier requires too much training data.
\item Identifying a robust classifier from limited training data is
  information theoretically possible but computationally intractable.
\item We just have not found the right algorithm yet.
\end{enumerate}
BPR provides evidence in favor of hypothesis 3 by constructing a binary classification task that admits a classifier robust to Euclidean perturbations of size $\log^{1/2 - \eps} n$ (while with high probability any sample has norm $O(\sqrt{n})$), yet finding {\em any} non-trivial robust classifier (even for arbitrarily small perturbations, and with probability of correctness only slightly better than chance) is hard in the statistical query model (we refer to BPR for a discussion of the statistical query model, including its weaknesses). In the present paper we significantly strengthen this result, while at the same time providing a much simpler proof. Our main contribution is to construct a binary classification task which admits a classifier robust to perturbations of size $\Omega(\sqrt{n})$ (i.e., a {\em maximally robust} classifier), yet finding {\em any} non-trivial robust classifier (even for arbitrarily small perturbations, and with probability of correctness only slightly better than chance) is impossible in polynomial time under a standard cryptographic assumption (namely, quadratic residuosity assumption).

We prove this new result in Section \ref{sec:proof}, where our main tool will be {\em trapdoor pseudorandom generators} (PRG). As a warm-up we start in Section \ref{sec:addendum} by providing a refinement of hypothesis 1 based on a simple observation with classical PRGs.

%% file: addendum.tex
\section{An addendum to hypothesis 1} \label{sec:addendum}
The basis of the present work is the observation that there is missing case in the four BPR hypotheses: the problem could admit a robust classifier $f$, which is ``easy to find" (say it is even given), but is intractable to compute (in the sense that the mapping $x \mapsto f(x)$ is hard to compute). To put it differently, a more appropriate version of hypothesis 1 would have been:
\begin{enumerate}
\item[1'.] No {\em efficiently computable} robust classifier exists.
\end{enumerate}
Let us now give a concrete construction that shows that hypothesis 1 and 1' are fundamentally different. Precisely we will exhibit a binary classification task that admits a maximally robust classifier, yet any {\em efficiently computable classifier} has an accuracy close to random guessing.

Let $G : \{0,1\}^{n/2} \rightarrow \{0,1\}^{n}$ be a PRG. Let $\cD_0$ be uniform on $\{0,1\}^n$ and $\cD_1$ be the distribution of $G(s)$ for $s$ uniform in $\{0,1\}^{n/2}$. Clearly a simple volume argument shows that there exists a classifier $A$ which satisfies \eqref{eq:robust} for $\epsilon = \Theta(\sqrt{n})$ (i.e., this problem admits a maximally robust classifier). Yet by definition of a PRG no polynomial time algorithm can have a non-trivial classification accuracy (let alone robust accuracy). 

%% file: proof.tex
\section{Adversarial examples and trapdoor PRG} \label{sec:proof}
Given the addendum from Section \ref{sec:addendum}, our goal is now to construct a classification task which admits a maximally robust classifier {\em that is also efficiently computable}, yet one cannot get non-trivial accuracy in polynomial time. The main idea here is to replace the PRG in the construction of Section \ref{sec:addendum} with a {\em trapdoor PRG}. In a nutshell a trapdoor PRG comes with a {\em key}, such that knowing the key allows to efficiently distinguish the PRG from a true source of randomness (and thus allows for efficient classification in the construction of Section \ref{sec:addendum}). Note also that, by a simple union bound, the sample complexity of such a problem would be of order of the number of bits in the key.

Let us now detail the construction a bit more. For the sake of concreteness, we use a specific trapdoor PRG, namely the Blum--Blum--Shub PRG \cite{BBS} (in its ``backward" form). Let $p$ and $q$ be large distinct prime numbers congruent to $3$ mod $4$, let $N = pq$ and $n= O(\log(N))$. The BBS PRG $G_{N}:\{0,1\}^n \rightarrow \{0,1\}^*$ works as follow. First it maps the seed $s \in \{0,1\}^n$ to $x_0 \in \N$ a quadratic residue mod $N$ in such a way that a uniformly random seed gives a nearly-uniform quadratic residue modulo $N$. Next it iteratively takes square roots mod $N$, that is let $x_{i+1}$ be such that $x_i = x_{i+1}^2 \mod N$ and $x_{i+1}$ is a quadratic residue itself (this is well-defined per our assumption on $p$ and $q$). The $i^{th}$ element of the output of $G_{N}$ is then simply the parity of $x_i$.

The key property of the BBS PRG is that, without knowing the factorization $N=pq$, its output is computationally indistinguishable (under the quadratic residuosity assumption) from a true source of randomness
(even when the seed is known), while on the other hand knowing the factorization allows for efficient distinguishing. To make this mathematically precise let us
recall the notion of \emph{computational} statistical distance for a family of pairs of distribution $\{(D_0(\omega), D_1(\omega)), \omega \in \Omega\}$: it is the supremum over all polynomial-time algorithms of the infimum over $\omega \in \Omega$ of the success probability one can have to identify whether a random sample was generated from $D_0(\omega)$ or generated from $D_1(\omega)$. Let us fix some constant $c>1$ and denote $\cD_0^n = \mathrm{unif}(\{0,1\}^{n^c})$ and $\cD_1^n(N)$ the distribution of the first $n^c$ bits of $s \circ G_N(s)$ where $s$ is a uniformly random element of $\{0,1\}^n$.
\begin{theorem}[\cite{BBS,VV,Gold}] \label{thm:BBS}
Assuming that for infinitely many $N$ the computational statistical distance of $\{(\cD_0^n, \cD_1^n(pq))\}_{p,q}$ is greater than $1/2 + 1/\mathrm{poly}(n)$
would refute the \emph{quadratic residuosity assumption}.

On the other hand, if $p$ and $q$ are known, then the computational statistical distance of $\{(\cD_0^n, \cD_1^n(N))\}$ is $1-o_n(1)$.
\end{theorem}

From the above discussion we have the following properties for the classification task described by the family $\{(\cD_0^n, \cD_1^n(pq))\}_{p,q}$:
\begin{enumerate}
\item[a.] The (robust) sample complexity of this family is $O(n)$.
\item[b.] Any task in this family admits a maximally robust classifier (same volume argument as in Section \ref{sec:addendum}) that is also efficiently computable (second statement in Theorem \ref{thm:BBS}).
\item[c.] Under the quadratic residuosity assumption, any polynomial time learning algorithm for this family has an accuracy close to chance on some task in the family (first statement in Theorem \ref{thm:BBS}).
\end{enumerate}
We also note that, using the BPR trick of adding a dummy coordinate revealing the label, one could replace property c by c' and add property d as follows (for any fixed $\epsilon>0$): 
\begin{enumerate}
\item[c'.] Under the quadratic residuosity assumption, any polynomial time learning algorithm for this family has a $\epsilon$-robust accuracy close to chance on some task in the family.
\item[d.] One can achieve \eqref{eq:nonrobust} in polynomial time (and polynomial sample complexity).
\end{enumerate}

%% file: main.bbl
\begin{thebibliography}{4}
\providecommand{\natexlab}[1]{#1}
\providecommand{\url}[1]{\texttt{#1}}
\expandafter\ifx\csname urlstyle\endcsname\relax
  \providecommand{\doi}[1]{doi: #1}\else
  \providecommand{\doi}{doi: \begingroup \urlstyle{rm}\Url}\fi

\bibitem[Blum et~al.(1986)Blum, Blum, and Shub]{BBS}
Lenore Blum, Manuel Blum, and Mike Shub.
\newblock A simple unpredictable pseudo-random number generator.
\newblock \emph{SIAM Journal on computing}, 15\penalty0 (2):\penalty0 364--383,
  1986.

\bibitem[Bubeck et~al.(2018)Bubeck, Price, and Razenshteyn]{BPR18}
S{\'e}bastien Bubeck, Eric Price, and Ilya Razenshteyn.
\newblock Adversarial examples from computational constraints, 2018.
\newblock URL \url{https://arxiv.org/abs/arXiv:1805.10204}.

\bibitem[Goldreich(2008)]{Gold}
Oded Goldreich.
\newblock Computational complexity: a conceptual perspective.
\newblock \emph{ACM Sigact News}, 39\penalty0 (3):\penalty0 35--39, 2008.

\bibitem[Vazirani and Vazirani(1983)]{VV}
Umesh~V Vazirani and Vijay~V Vazirani.
\newblock Trapdoor pseudo-random number generators, with applications to
  protocol design.
\newblock In \emph{Foundations of Computer Science, 1983., 24th Annual
  Symposium on}, pages 23--30. IEEE, 1983.

\end{thebibliography}
